\documentclass[letterpaper]{article} 
\usepackage{aaai23}  
\usepackage{times}  
\usepackage{helvet}  
\usepackage{courier}  
\usepackage[hyphens]{url}  
\usepackage{graphicx} 
\usepackage{natbib}  
\usepackage{caption} 
\usepackage{algorithm}
\usepackage{algorithmic}

\usepackage{booktabs}       
\usepackage{amsfonts}       
\usepackage{nicefrac}       
\usepackage{microtype}      
\usepackage{xcolor}         

\usepackage{verbatim}
\usepackage{amsmath}
\usepackage{graphicx,psfrag}
\usepackage{enumerate}
\newcommand{\R}{\mathbb{R}}
\usepackage{newfloat}
\usepackage{listings}
\DeclareCaptionStyle{ruled}{labelfont=normalfont,labelsep=colon,strut=off} 
\floatstyle{ruled}
\newfloat{listing}{tb}{lst}{}
\floatname{listing}{Listing}
\title{Online Detection Of Supply Chain Network Disruptions Using Sequential Change-Point Detection for Hawkes Processes}
\author {
    Khurram Yamin,\textsuperscript{\rm 1}
    Haoyun Wang, \textsuperscript{\rm 1}
    Benoit Montreuil ,\textsuperscript{\rm 1}
    Yao Xie \textsuperscript{\rm 1}
}
\affiliations {
    \textsuperscript{\rm 1} Georgia Tech, Department of Industrial and Systems Engineering \\
    kyamin3@gatech.edu, hwang800@gatech.edu, benoit.montreuil@isye.gatech.edu,yao.xie@isye.gatech.edu
}

\usepackage{bibentry}
\nocopyright
\begin{document}

\maketitle

\frenchspacing
\begin{abstract}
In this paper, we attempt to detect an inflection or change-point resulting from the Covid-19 pandemic on supply chain data received from a large furniture company. To accomplish this, we utilize a modified CUSUM (Cumulative Sum) procedure on the company's spatial-temporal order data as well as a GLR (Generalized Likelihood Ratio) based method. We model the order data using the Hawkes Process Network, a multi-dimensional self and mutually exciting point process, by discretizing the spatial data and treating each order as an event that has a corresponding node and time. We apply the methodologies on the company's most ordered item on a national scale and perform a deep dive into a single state. Because the item was ordered infrequently in the state compared to the nation, this approach allows us to show efficacy upon different degrees of data sparsity. Furthermore, it showcases use potential across differing levels of spatial detail.
\end{abstract}

\section{Introduction}

Covid-19 caused wide sweeping changes in society and affected all of the elements of our lives from the way we work to the way we socialize. As such, it created new needs, and it is logical to question whether supply chains were affected as people looked for new products to satisfy those needs. Such a question is normally extremely difficult to answer because of the lack of publicly available, up to date, and robust supply chain data. We, however, are in the fortunate position of having access to some of the proprietary data of a large furniture company that operates internationally. 

We choose to model the network of orders as a point process because the relative scarcity of the data limits the use of some traditional time series models (e.g. auto-regressive process). Specifically, we use a Hawkes process network, a type of point process which is mutually and self exciting, where each node represents a territorial region such as a state or county depending on the level of analysis. The rationale behind the use of the Hawkes process is that it has an element that accounts for a triggering influence of past events on future ones. For example, if a customer buys an item of furniture and likes it, it is possible that she recommends it to a friend. 

To detect change-points resulting from Covid-19, we apply two methods: a modified CUSUM (Cumulative Sum) procedure proposed by \cite{wang_xie_xie_cuozzo_mak_2022} and the window-limited GLR (Generalized Likelihood Ratio) based procedure discussed in \cite{li2017detecting}. The two method serves different scenarios. When we have a reasonable estimate of the form of the inflection, CUSUM is computation efficient and has better performance. The GLR on the other hand can deal with completely unexpected anomaly types.

To our knowledge, this is the first attempt at using sequential change-point detection algorithms on real supply chain data in a peer-reviewed setting. As such, it serves as a proof of concept and a potential framework for others who wish to try a similar task. We believe the use of the Hawkes process network could be applicable to many other supply chain situations given a similar triggering effect would be present. Additionally, we show that the modified CUSUM method is not only effective in simulation, but in a real life case. 

\section{Literature Review}
The Hawkes process was proposed in \cite{10.2307/2334319} and has since become popular for use in modeling situations where there is a clear triggering effect. For example, it has been used in modeling for seismology in \cite{ogata_1998} as an earthquake is generally followed by aftershocks. It has also been used to model biological neural networks in \cite{6736879} as the firing of other neurons has an impact on the firing other neurons in the network. Other applications include stock price \citep{embrechts2011multivariate}, crime events \citep{mohler2013modeling}, social media activities \citep{rizoiu2017tutorial}.

However, there has been far less work done on change-point detection for the Hawkes process. In addition, the majority of the work that has been done has explored offline change-point detection which references the entire time series and do not work in real time. For example, \cite{rambaldi} develops a procedure for financial data to find the times when bursts in the intensity of the Hawkes process begin. However in our paper, we seek to detect change-points in real time as quickly as possible using an online method. In terms of such work, \cite{7907333} examines the use of the GLR (Generalized Likelihood Ratio) test for Hawkes processes in situations in which the post-change parameters were not known. \cite{wang_xie_xie_cuozzo_mak_2022} develops a memory efficient modified CUSUM implementation that gives less consideration to events in the far past. This algorithm is further explained in the Background Section. The previously mentioned GLR and modified CUSUM algorithms are two tests we use on our supply chain data.

\section{Background and Problem Formulation}
\label{sec:background}
In this section we describe the Hawkes process model on networks as well as the change-point detection problem, and how the two detecting procedures are applied to the furniture sales data.
\subsection{Hawkes Process}
The Hawkes process is one kind of point process, which models the probability of events happening in continuous time. It can be characterized by the intensity function $\lambda:\R^+\to\R^+$, where at each time $t\geq 0$ the intensity $\lambda(t)$ is the probability that a new event happens in the infinitesimal near future, $$
\mathbb P(\text{new event in }(t,t+dt)) = \lambda(t) dt.
$$
An example is the Poisson process, where the intensity function is a constant, $\lambda(t) = \mu, \forall t\geq 0.$ In the Hawkes process, the intensity $\lambda(t)$ is decided by the history. Let $0<t_1<t_2<t_3<\dots$ be the events occurrence times. Let $\mathcal H_t$ be the set of event occurrence times up to and including time $t$, and $N_t$ be counting process, i.e. the number of events in $\mathcal H_t$. The Hawkes process captures the triggering effect between events by letting 
$$
\lambda(t) = \mu + \int_0^t \alpha\varphi(t - \tau)dN_\tau,
$$
where $\mu> 0$ is the background intensity, $\varphi$ is a non-negative kernel function describing how the influence of an event is distributed into the future. A common choice is $\varphi(t) = \beta\exp(-\beta t)$ for some $\beta$. $\alpha\geq 0$ is the magnitude of the triggering effect. The integral over the counting measure $N_t$ is equivalent to taking the summation over past event times.

\textbf{Hawkes process with marks.} The Hawkes process can be generalized to fit events data with marks to capture the triggering effect acted through information other than temporal relationship. In our setup, each event $i$ is an order placed online by individual customers, with the occurrence time $t_i$ and the shipping address $u_i$ discretized into states or zip codes. Then the orders can be treated as event data on a network, where each node is one location. For each pair of nodes with index $i$ and $j$, we consider the triggering effect $\alpha_{ij}$ from $i$ to $j$ when the two geographical locations are adjacent. Then for each node $i$, the intensity function $\lambda_i(t)$ at this location is
$$
\lambda_i(t) = \mu_i + \sum_{j=1}^D\int_0^t \alpha_{ij}\varphi(t-\tau)dN_\tau^j,
$$
here $D$ is the size of the network, $\mu_i> 0,N_t^i$ is the background intensity and counting process on node $i$. The kernel function $\varphi$ is exponential.

\subsection{Change-point Detection}
We look to detect an abrupt change in the parameters $(\mu_i)_{i=1}^D$, $(\alpha_{ij})_{i,j=1}^D$ as quickly as possible. The problem can be formulated as follows:
\begin{align*}
    H_0:&\ \lambda_i(t) = \mu_{i,0} + \sum_{j=1}^D\int_0^t \alpha_{ij,0}\varphi(t-\tau)dN_\tau^j,~~t\geq 0\\
    H_1:&\ \lambda_i(t) = \mu_{i,0} + \sum_{j=1}^D\int_0^t \alpha_{ij,0}\varphi(t-\tau)dN_\tau^j,~~0\leq t\leq  \kappa\\
    &\ \lambda_i(t) = \mu_{i,1} + \sum_{j=1}^D\int_\kappa^t \alpha_{ij,1}\varphi(t-\tau)dN_\tau^j,~~t> \kappa.
\end{align*}
Here $\kappa$ is the unknown change-point. Then we carry out a repeated test to decide whether there has been a change-point. The pre-change model is learnt using historical data, including the decay rate $\beta$ in the kernel function $\varphi$. The post-change parameters $(\mu_{i,1})_{i=1}^D, (\alpha_{ij,1})_{i,j=1}^D$ sometimes represent an unexpected anomaly and thus are treated as unknown. In correspondence to whether or not the post-change parameters are known or can be estimated, we apply the CUSUM and GLR procedure described in the following paragraphs.

\textbf{CUSUM} \citep{wang_xie_xie_cuozzo_mak_2022}. When the post-change parameters can be estimated accurately, CUSUM is a computational and memory efficient detecting procedure which also enjoys asymptotic optimality in performance. Similar with the traditional i.i.d. case, here the CUSUM statistic over the dynamic Hawkes network is
$$
S_t^{\rm CUSUM} = \sup_{0\leq \nu\leq t} \ell_{\nu,t},
$$
where $\ell_{\nu,t}$ is the log-likelihood ratio up to time $t$ between $H_1$ and $H_0$ as if $\nu$ is the true change-point,
$$
\ell_{\nu,t} = \sum_{i=1}^D \int_{\nu}^t (\lambda_{i,\infty}(\tau) - \lambda_{i,\nu}(\tau))(dN_\tau^i-d\tau),
$$
where $\lambda_{i,\nu}(t)$ is the intensity at node $i$ as if $\nu$ is the true change-point, and we use infinity for the case under $H_0$. The procedure raises an alarm when the statistic $S_t^{\rm CUSUM}$ exceeds some pre-determined threshold (same with GLR). With a proper truncation on the kernel function $\varphi$, the CUSUM statistic can be computed recursively with high precision. Also the integral can be replaced with the sum over past events to avoid numerical evaluation because we know the closed form expression of $\int_0^t\varphi(\tau)d\tau$ for every $t$.

\textbf{GLR} \citep{li2017detecting}. When the post-change parameters are unknown, we can compute the generalized likelihood ratio in a sliding window to reflect the difference between the current data and the pre-change model. For a properly chosen window length $w$ that we believe is long enough to successfully capture the change and yet not too large which results in a large detection delay, the GLR statistic is
$$
S_t^{\rm GLR} = \sup_{\boldsymbol\mu_1,A_1} \ell_{t-w,t,\boldsymbol\mu_1,A_1},
$$
where $\ell_{t-w,t,\boldsymbol\mu_1,A_1}$ is the log-likelihood ratio up to time $t$ as if $t-w$ is the true change-point, $\boldsymbol\mu_1=(\mu_i)_{i=1}^D,A_1 = (\alpha_{ij})_{i,j=1}^D$ are the post-change parameters. For each $t$, $\ell_{t-w,t,\boldsymbol\mu_1,A_1}$ is convex in $\boldsymbol\mu_1,A_1$, and the supremum can be found using the Expectation-Maximization (EM) algorithm. The merit of the GLR procedure is it can detect an unexpected and unknown change, while also gives the estimated post-change parameters when it raises an alarm.

\section{Experimental Setup}
In each of the experiments, we use the March 2018 to March 2019 data to train the pre-change parameters of the Hawkes network using the likelihood function. To verify the effectiveness of the change-point detection procedures, we would expect the statistics $S_t^{\rm CUSUM},S_t^{\rm GLR}$ to remain small until roughly March 2020 when the WHO declared Covid-19 a global pandemic \cite{katella_2021}, and raise significantly after that. The decay rate $1/\beta = 5$ days in the kernel function $\varphi$ is found manually since the likelihood ratio is non-convex over $\beta$, and the pre-change parameters $\boldsymbol\mu_0 = (\mu_{i,0})_{i=1}^D, A_0 = (\alpha_{ij,0})_{i,j=1}^D$ are the maximum likelihood estimates (MLE). For the GLR, we design a window length $w = 100$ days based on life experiences. For the CUSUM, the post-change parameter $\mu_1$ can be set to $2\mu_0$ or $0.5\mu_0$ to detect a change in average demand. If it is desired to detect a local change in the demand correlation, we can design the post-change $A_1$ to have vanishing edges especially for large states such as California/Texas/Florida/Pennsylvania. We believe more meaningful post-change parameters can be designed with a better understanding of the reasons behind the average demand and spatial correlation. 

\section{Work Desk - United States}

We analyze the sales of a specific work desk which is the mostly commonly ordered item in the company on the national level. We group the addresses into the 50 states and Washington DC for the purpose of creating nodes in the Hawkes process. There is an average of 40 orders per day over the 50 states and Washington DC. You can see from Figure 1, there is a drastic jump in orders between the start and end of March 2020. 

\begin{figure}[htb]

\begin{minipage}[b]{.48\linewidth}
  \centering
  \centerline{\includegraphics[width=3.5cm]{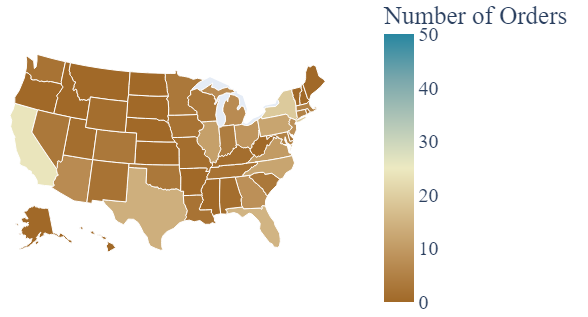}}
  \centerline{(c)}\medskip
\end{minipage}
\hfill
\begin{minipage}[b]{0.48\linewidth}
  \centering
  \centerline{\includegraphics[width=3.5cm]{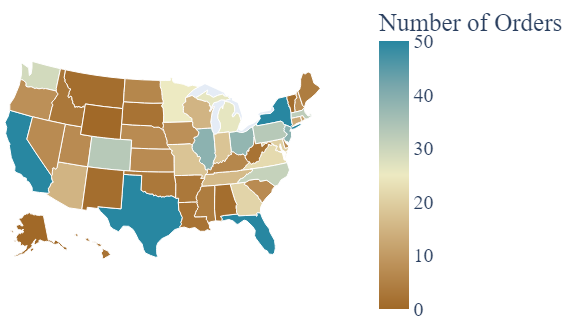}}
  \centerline{(d)}\medskip
\end{minipage}
\caption{A mapping of the number of orders in each state for a week, for the weeks of (a) 2020-02-23, (b) 2020-04-12}
\label{fig:south_shore_modelo}
\end{figure}

\subsection{Detection with GLR}
As can be seen from Figure 2(a), the GLR score spikes after March of 2020 in a way that it never does between March of 2019 and March of 2020. As shown in Figure 3(a) which displays the fitted pre-change Hawkes process model, there exists a very strong causal effect between states, especially from the more populated states to their neighboring ones. Additionally we can see in Figure 3(b) which models the fitted post-change Hawkes process that when the detecting procedure raises an alarm at the beginning of Covid, the most salient change in the fitted model is in the magnitude of the background intensities (represented as node sizes), as well as the disappearing influence from California to its neighboring states.

\begin{figure}[htb]

\begin{minipage}[b]{.48\linewidth}
  \centering
  \centerline{\includegraphics[width=4.6cm,height=3.6cm]{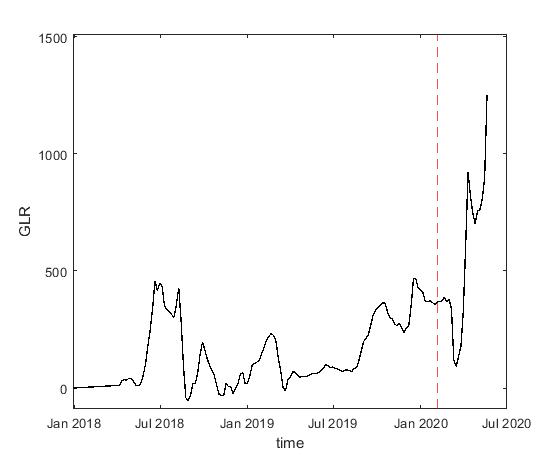}}
  \centerline{(a)}\medskip
\end{minipage}
\hfill
\begin{minipage}[b]{.48\linewidth}
  \centering
  \centerline{\includegraphics[width=4.6cm]{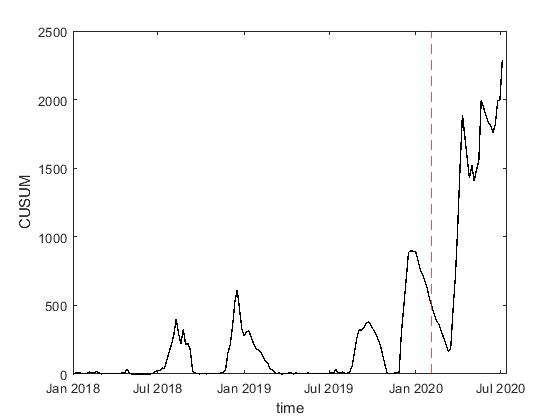}}
  \centerline{(b)}\medskip
\end{minipage}

\caption{(a) GLR and (b) CUSUM statistic over time for national orders. The x-axis is in days starting from January 21st, 2018. The vertical line marks March 1st, 2020, when Covid was declared a pandemic.}
\label{fig:southshore_stat}
\end{figure}

\begin{figure}[htb]

\begin{minipage}[b]{0.48\linewidth}
  \centering
  \centerline{\includegraphics[width=4cm]{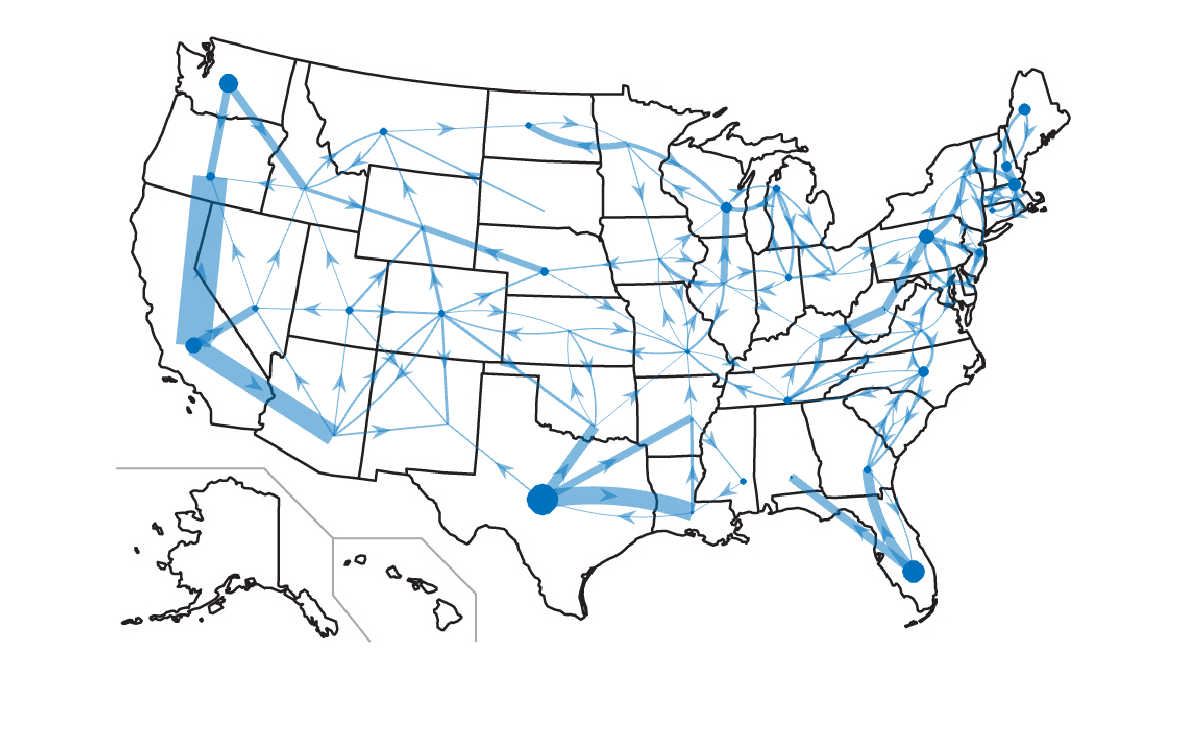}}
  \centerline{(a)}\medskip
\end{minipage}
\hfill
\begin{minipage}[b]{0.48\linewidth}
  \centering
  \centerline{\includegraphics[width=4cm]{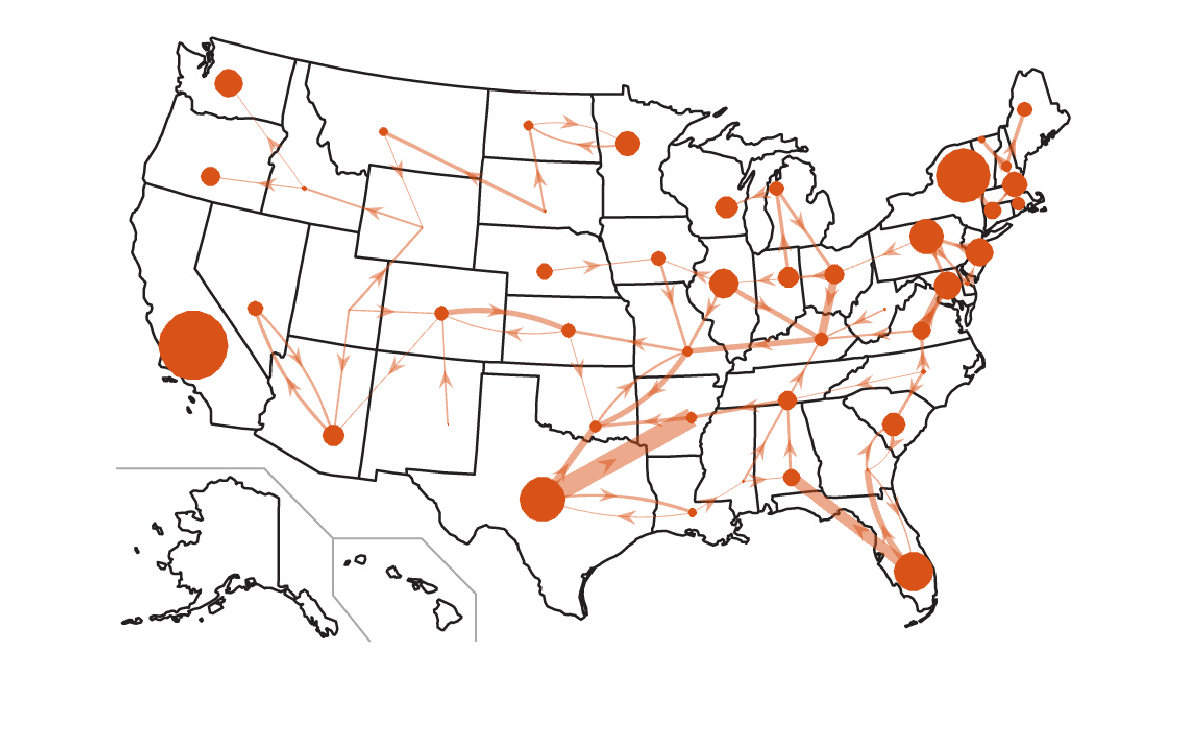}}
  \centerline{(b)}\medskip
\end{minipage}
\caption{(a) The fitted pre-change model and (b) the fitted model at the beginning of Covid when GLR raises an alarm. The width of the directed edges corresponds to the interstate influences, while the size of the node is proportional to the background intensity.} 
\label{fig:south_shore_model}
\end{figure}

\subsection{Detection with CUSUM}

We tested both possible post-change parameter for $\mu_1$, $2\mu_0$ and $0.5\mu_0$. As shown in Figure 2 (b), we were able to successfully capture the disruption in the distribution of orders caused by Covid-19 by doubling $\mu_0$. Given the jump in demand in April 2020 that can be observed in Figure 1, it would make sense in hindsight to attempt to detect a surge in demand. The spike in CUSUM after March 2020 is clearly visible and greatly differs in magnitude in comparison to any of the smaller spikes preceding it. CUSUM appears to contain slightly less noise but both the CUSUM and GLR are very comparable in this situation. For CUSUM, we don't perform an analysis of pre-change vs post-change parameters as post-change parameters are predetermined.

\section{Work Desk-California}
For experimentation with finer granularity, we isolate the subset of the orders that came from California. We then group the addresses into the 54 counties that exist in the state for which we have data to use for the Hawkes Process nodes. There is an average of 3.5 orders a day, meaning the data is more sparse than the national level orders. The counties with the biggest increase in orders are all in the densely populated Southern part of the state as can be seen from Figure 4.

\begin{figure}[htb]

\begin{minipage}[b]{.48\linewidth}
  \centering
  \centerline{\includegraphics[width=3.6cm]{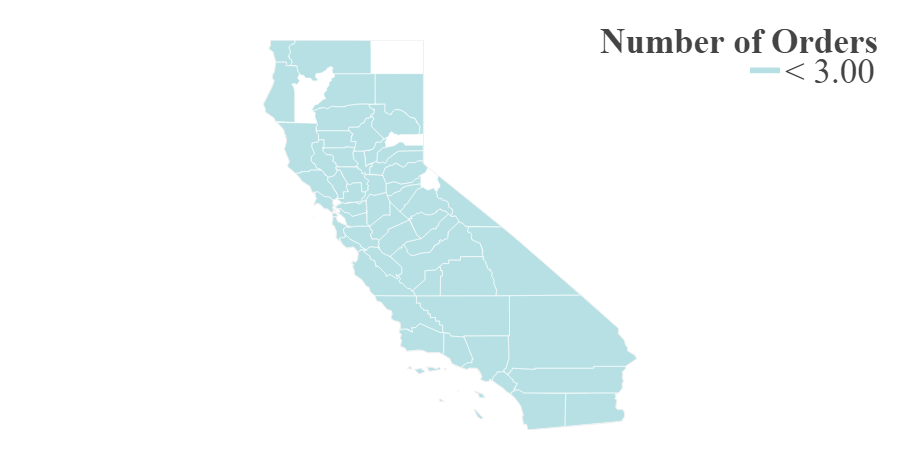}}
  \centerline{(a)}\medskip
\end{minipage}
\hfill
\begin{minipage}[b]{0.48\linewidth}
  \centering
  \centerline{\includegraphics[width=3.6cm]{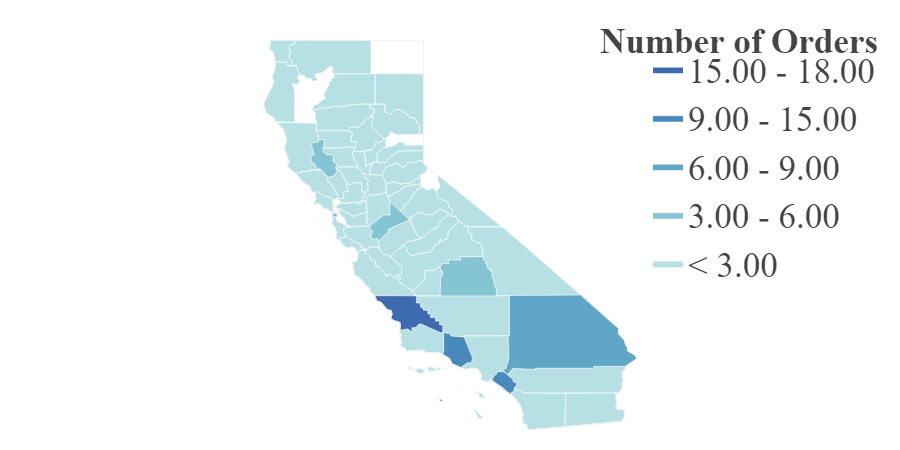}}
  \centerline{(b)}\medskip
\end{minipage}
\caption{A mapping of the number of orders in each county in California for the weeks of (a) 2020-02-23, (b) 2020-04-12} 
    \label{fig:south_shore_modelo_cali}

\end{figure}

\subsection{Detection with GLR}
As can be noted in Figure 5 (a), GLR spikes post March 2020 in a way that exceeds the previous apex of the GLR score. However, this spike is not extreme in magnitude when compared to the spikes that came between March 2019 and March 2020, or to the relative magnitude of the post March 2020 spike that we saw in the GLR score of the national case. There are some interesting patterns in the pre-change fitted Hawkes process displayed in Figure 6(a).  Several of the counties in the middle part of California have small populations but still exhibit some influence on surrounding counties possibly because of how close the residents of those counties are in proximity. In the South, we see stronger influences from more populous counties. Moving to Figure 6(b), we see major shifts in inter-county influences in Southern California.

\begin{figure}[htb]

\begin{minipage}[b]{.48\linewidth}
  \centering
  \centerline{\includegraphics[width=4.6cm]{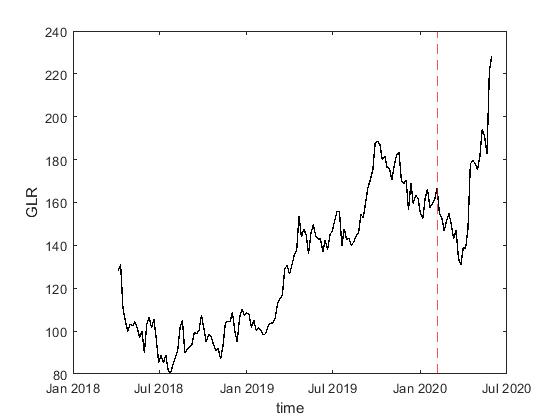}}
  \centerline{(a)}\medskip
\end{minipage}
\hfill
\begin{minipage}[b]{.48\linewidth}
  \centering
  \centerline{\includegraphics[width=4.6cm]{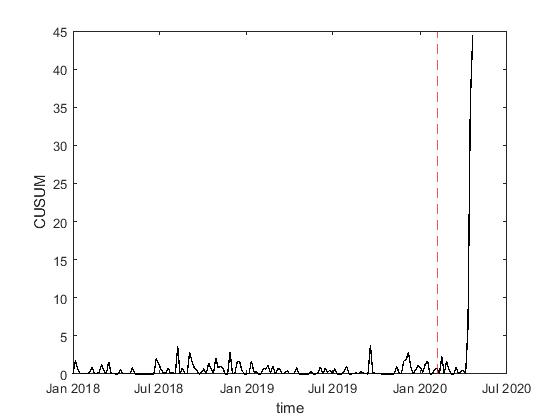}}
  \centerline{(b)}\medskip
\end{minipage}

\label{fig:southshore_stat_cali}
\caption{GLR and CUSUM statistic over time for California orders. The x-axis is in days starting from January 21st, 2018. The vertical line marks March 1st, 2020.}
\end{figure}

\begin{figure}[htb]

\begin{minipage}[b]{0.4\linewidth}
  \centering
  \centerline{\includegraphics[width=2.8cm]{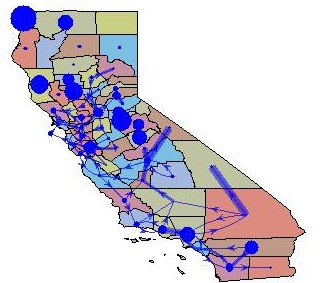}}
  \centerline{(a)}\medskip
\end{minipage}
\hfill
\begin{minipage}[b]{0.4\linewidth}
  \centering
  \centerline{\includegraphics[width=2.8cm]{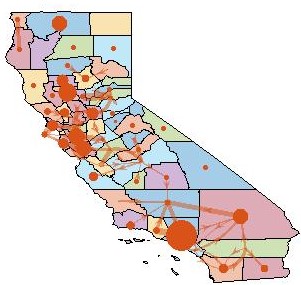}}
  \centerline{(b)}\medskip
\end{minipage}
 \caption{(a)The fitted pre-change model and (b) post-change model. The width of the directed edges corresponds to the intercounty influences. County color coding is not significant. } 
    \label{fig:south_shore_model_cali}
\end{figure}

\subsection{Detection with CUSUM}
We again test the possible post-change parameters design and ended up finding success in setting the post-change $\mu_1=2\mu_0$. The CUSUM model is then successfully able to detect the surge in demand as can be observed in Figure 5(b). In hindsight, given what we can see in Figure 4, specifically the jump in demand in South California, this formulation makes sense. The model has an extreme spike in CUSUM score after the beginning of Covid while the preceding year is relatively flat. In such a way, the change point is far more clearly defined in the CUSUM model than the GLR model.  

\section{Discussion}
Sequential change-point detection is a valuable tool for monitoring inflections in temporal data. This is the first time it has been applied to real supply chain data in peer reviewed literature. There are several novel features regarding the techniques used. Firstly, we show that the Hawkes Process can successfully be used to model a specific supply chain network. This has evident implications for other situations in which there is a clear triggering effect and data sparsity precludes the use of other traditional methods.  

Secondly, we show that the GLR and CUSUM procedure can be successfully applied to online detection of change-points in the supply chain. Both methodologies were able to successfully detect surges in demand as they were happening. However, CUSUM performed far better than GLR in the case of California and exhibited far less noise. This may have been because the specification of reasonable post change parameters prior to running the CUSUM algorithm.

There are obviously however several limitations to the approach that we used. Firstly, under regular conditions, it is not an easy task to verify the change points that have been detected. The algorithm detects a signal but it may be difficult to understand what that signal corresponds to. Additionally, a limitation of the CUSUM algorithm in particular is that it requires post-change parameter specification. This presents challenges in situations where post change parameters are difficult to predict. 
In future work, it would be interesting to apply change-point detection on other tiers of supply chain  data. For example the furniture company also makes sales to downstream sellers such as Amazon and Wayfair. Such orders comes in batches and happen less frequently than sales made to individual buyers, which makes the fitting and change-point detection potentially harder.

\bibliography{aaai23}

\begin{thebibliography}{11}
\providecommand{\natexlab}[1]{#1}

\bibitem[{Embrechts, Liniger, and Lin(2011)}]{embrechts2011multivariate}
Embrechts, P.; Liniger, T.; and Lin, L. 2011.
\newblock Multivariate Hawkes processes: an application to financial data.
\newblock \emph{Journal of Applied Probability}, 48(A): 367--378.

\bibitem[{Hawkes(1971)}]{10.2307/2334319}
Hawkes, A.~G. 1971.
\newblock Spectra of Some Self-Exciting and Mutually Exciting Point Processes.
\newblock \emph{Biometrika}, 58(1): 83--90.

\bibitem[{Katella(2021)}]{katella_2021}
Katella, K. 2021.
\newblock Our pandemic year-A covid-19 timeline.

\bibitem[{Li et~al.(2017{\natexlab{a}})Li, Xie, Farajtabar, Verma, and
  Song}]{li2017detecting}
Li, S.; Xie, Y.; Farajtabar, M.; Verma, A.; and Song, L. 2017{\natexlab{a}}.
\newblock Detecting changes in dynamic events over networks.
\newblock \emph{IEEE Transactions on Signal and Information Processing over
  Networks}, 3(2): 346--359.

\bibitem[{Li et~al.(2017{\natexlab{b}})Li, Xie, Farajtabar, Verma, and
  Song}]{7907333}
Li, S.; Xie, Y.; Farajtabar, M.; Verma, A.; and Song, L. 2017{\natexlab{b}}.
\newblock Detecting Changes in Dynamic Events Over Networks.
\newblock \emph{IEEE Transactions on Signal and Information Processing over
  Networks}, 3(2): 346--359.

\bibitem[{Mohler(2013)}]{mohler2013modeling}
Mohler, G. 2013.
\newblock Modeling and estimation of multi-source clustering in crime and
  security data.
\newblock \emph{The Annals of Applied Statistics}, 1525--1539.

\bibitem[{Ogata(1998)}]{ogata_1998}
Ogata, Y. 1998.
\newblock Space-time point-process models for earthquake occurrences.
\newblock \emph{Annals of the Institute of Statistical Mathematics}, 50(2):
  379–402.

\bibitem[{Rambaldi, Filimonov, and Lillo(2016)}]{rambaldi}
Rambaldi, M.; Filimonov, V.; and Lillo, F. 2016.
\newblock Detection of intensity bursts using Hawkes processes: An application
  to high-frequency financial data.
\newblock \emph{Physical Review E}, 97.

\bibitem[{Reynaud-Bouret, Rivoirard, and Tuleau-Malot(2013)}]{6736879}
Reynaud-Bouret, P.; Rivoirard, V.; and Tuleau-Malot, C. 2013.
\newblock Inference of functional connectivity in Neurosciences via Hawkes
  processes.
\newblock In \emph{2013 IEEE Global Conference on Signal and Information
  Processing}, 317--320.

\bibitem[{Rizoiu et~al.(2017)Rizoiu, Lee, Mishra, and Xie}]{rizoiu2017tutorial}
Rizoiu, M.-A.; Lee, Y.; Mishra, S.; and Xie, L. 2017.
\newblock A tutorial on hawkes processes for events in social media.
\newblock \emph{arXiv preprint arXiv:1708.06401}.

\bibitem[{Wang et~al.(2022)Wang, Xie, Xie, Cuozzo, and
  Mak}]{wang_xie_xie_cuozzo_mak_2022}
Wang, H.; Xie, L.; Xie, Y.; Cuozzo, A.; and Mak, S. 2022.
\newblock Sequential change-point detection for mutually exciting point
  processes.
\newblock \emph{Technometrics}, 1–13.

\end{thebibliography}

\end{document}